\documentclass[runningheads]{llncs}
\usepackage{graphicx}
\usepackage{xcolor}
\usepackage{fancyvrb}
\usepackage{url}
\begin{document}
\title{OBA: An Ontology-Based Framework for Creating REST APIs for Knowledge Graphs}
\titlerunning{The Ontology Based API Framework (OBA)}
%
\author{Daniel Garijo \and
Maximiliano Osorio }
\authorrunning{Daniel Garijo and Maximiliano Osorio}
%
\institute{Information Sciences Institute, University of Southern California
\email{\{dgarijo,mosorio,gil\}@isi.edu}}
\maketitle              
\begin{abstract}
In recent years, Semantic Web technologies have been increasingly adopted by researchers, industry and public institutions to describe and link data on the Web, create web annotations and consume large knowledge graphs like Wikidata and DBPedia. However, there is still a knowledge gap between ontology engineers, who design, populate and create knowledge graphs; and web developers, who need to understand, access and query these knowledge graphs but are not familiar with ontologies, RDF or SPARQL. In this paper we describe the Ontology-Based APIs framework (OBA), our approach to automatically create REST APIs from ontologies while following RESTful API best practices. Given an ontology (or ontology network) OBA uses standard technologies familiar to web developers (OpenAPI Specification, JSON) and combines them with W3C standards (OWL, JSON-LD frames and SPARQL) to create maintainable APIs with documentation, units tests, automated validation of resources and clients (in Python, Javascript, etc.) for non Semantic Web experts to access the contents of a target knowledge graph. We showcase OBA with three examples that illustrate the capabilities of the framework for different ontologies.

\textbf{Resource type}: Software\\
\textbf{License}: Apache 2.0\\
\textbf{DOI}: https://doi.org/10.5281/zenodo.3686266\\
\textbf{Repository}: https://github.com/KnowledgeCaptureAndDiscovery/OBA/

\end{abstract}
\section{Introduction}

Knowledge graphs have become a popular technology for representing structured information on the Web. The Linked Open Data Cloud\footnote{\url{https://lod-cloud.net/}} contains more than 1200 linked knowledge graphs contributed by researchers and public institutions. Major  companies like Google,\footnote{\url{https://developers.google.com/knowledge-graph}} Microsoft,\footnote{\url{https://www.microsoft.com/en-us/research/project/microsoft-academic-graph/}} or Amazon \cite{amazon}  use knowledge graphs to represent some of their information. Recently, crowdsourced knowledge graphs such as Wikidata \cite{vrandecic_wikidata:_2014} have surpassed Wikipedia in the number of contributions made by users.


In order to create and structure these knowledge graphs, ontology engineers develop vocabularies and ontologies defining the semantics of the classes, object properties and data properties represented in the data. These ontologies are then used in  extraction-transform-load pipelines to populate knowledge graphs with data and make the result accessible on the Web to be queried by users (usually as an RDF dump or a SPARQL endpoint). However, consuming and contributing to knowledge graphs exposed in this manner is problematic for two main reasons. First, exploring and using the contents of a knowledge graph is a time consuming task, even for experienced ontology engineers (common problems include lack of usage examples that indicate how to retrieve resources, the ontologies used are not properly documented or without examples, the format in which the results are returned is hard to manipulate, etc.). Second, W3C standards such as SPARQL \cite{Seaborne:13:SQL} and RDF \cite{Cyganiak:14:RCA} are still unknown to a major portion of the web developer community (used to JSON and REST APIs), making it difficult for them to use knowledge graphs even when documentation is available. 


In this paper we address these problems by introducing OBA, an Ontology-Based API framework that given an ontology (or ontology network) as input, creates a JSON-based REST API server that is consistent with the classes and properties in the ontology; and can be configured to retrieve, edit, add or delete resources from a knowledge graph. OBA's contributions include:

\begin{itemize}
    \item A method for automatically creating a \textbf{documented REST OpenAPI specification}\footnote{\url{http://spec.openapis.org/oas/v3.0.3}} from an OWL ontology \cite{Patel-Schneider:12:OWO}, together with the means to customize it as needed (e.g., filtering some of its classes). Using OBA, new changes made to an ontology can be automatically propagated to the corresponding API, making it easier to maintain.
    \item A framework to create a \textbf{server implementation} based on the API specification to handle requests automatically against a target knowledge graph. The implementation will validate posted resources to the API and will deliver the results in a JSON format as defined in the API specification. 
    \item A method for \textbf{converting JSON-LD} returned by a SPARQL endpoint \cite{Kellogg:20:JSONLD} into JSON according to the format defined in the API specification.
    \item A mechanism based on named graphs\footnote{\url{https://www.w3.org/2004/03/trix/}} for users to \textbf{contribute} to a knowledge graph through POST requests.
    \item Automatic generation of \textbf{tests} for API  validation against a knowledge graph.
\end{itemize}

OBA uses W3C standards widely used in Web development (JSON) for accepting requests and returning results, while using SPARQL and JSON-LD frames to query knowledge graphs and frame data in JSON-LD. We consider that OBA is a valuable resource for the community, as it helps bridging the  gap between ontology engineers who design and populate knowledge graphs and application and service developers who can benefit from them. 

The rest of the paper is structured as follows: Section \ref{sec:relWork} discusses efforts from the Semantic Web community to help developers access knowledge graphs, while Section \ref{sec:oba} describes the architecture and rationale of the OBA framework. Section \ref{sec:showcase} shows the different features of OBA through three different examples, Section \ref{sec:adoption} discusses adoption, potential impact and current limitations of the tool, and Section \ref{sec:conclusions} concludes the paper.

\section{Related Work}\label{sec:relWork}
The Semantic Web community has developed different approaches for helping developers access and manipulate the contents of knowledge graphs. On the one hand, some approaches rely on mechanisms that are familiar to developers for helping them consume Linked Data. For instance, the W3C Linked Platform \cite{Malhotra:15:LDP} proposes a platform for handling HTTP requests over RDF data using containers for HTTP operations. Closer to our scope, Basil \cite{basil}, GRLC \cite{sack_grlc_2016} and r4r\footnote{\url{https://github.com/oeg-upm/r4r}} propose to create REST APIs from SPARQL queries for accessing knowledge graphs. The limitations of these approaches are that knowledge engineers have to define the queries that need to be supported by hand. In addition, API paths are edited manually and, as a result, the resultant REST API specifications do not often follow the standard practices promoted by the OpenAPI specification.

On the other hand, researchers have attempted to improve the serialization of SPARQL results. For example, SPARQL transformer \cite{sparqltransformer} and SPARQL-JSONLD\footnote{\url{https://github.com/usc-isi-i2/sparql-jsonld}} both present approaches for transforming SPARQL to user-friendly JSON results by using a custom mapping language and JSON-LD frames \cite{Kellogg:20:JSONLD} respectively. In \cite{Taelman2018GraphQLLDLD} the authors use  GraphQL,\footnote{\url{https://graphql.org/}} which is gaining popularity among the developer community, to generate SPARQL queries and serialize the results in JSON. While these approaches share in their aim the ability to facilitate obtaining JSON from a knowledge graph, developers still need to be familiar with the underlying ontologies used to query the data in those knowledge graphs.

Finally, \cite{rest_owlapi} proposes to define REST APIs to access the classes and properties of an ontology. This is different from our scope, which uses the ontology as a template to create an API to exploit the contents of a knowledge graph. To the best of our knowledge, our work is the first end-to-end framework for creating REST APIs from OWL ontologies to provide access to the contents of a knowledge graph.



\section{The Ontology Based APIs Framework (OBA)}\label{sec:oba} 
OBA is a framework designed to help ontology engineers create RESTful APIs from ontologies. Given an OWL ontology or ontology network and a knowledge graph (accessible through a SPARQL endpoint), OBA automatically generates a documented standard API specification and creates a REST API server that can validate requests from users, test all API calls and deliver JSON objects following the structure described in the ontology. 

\begin{figure}[t!]
\centering
\includegraphics[width=\textwidth]{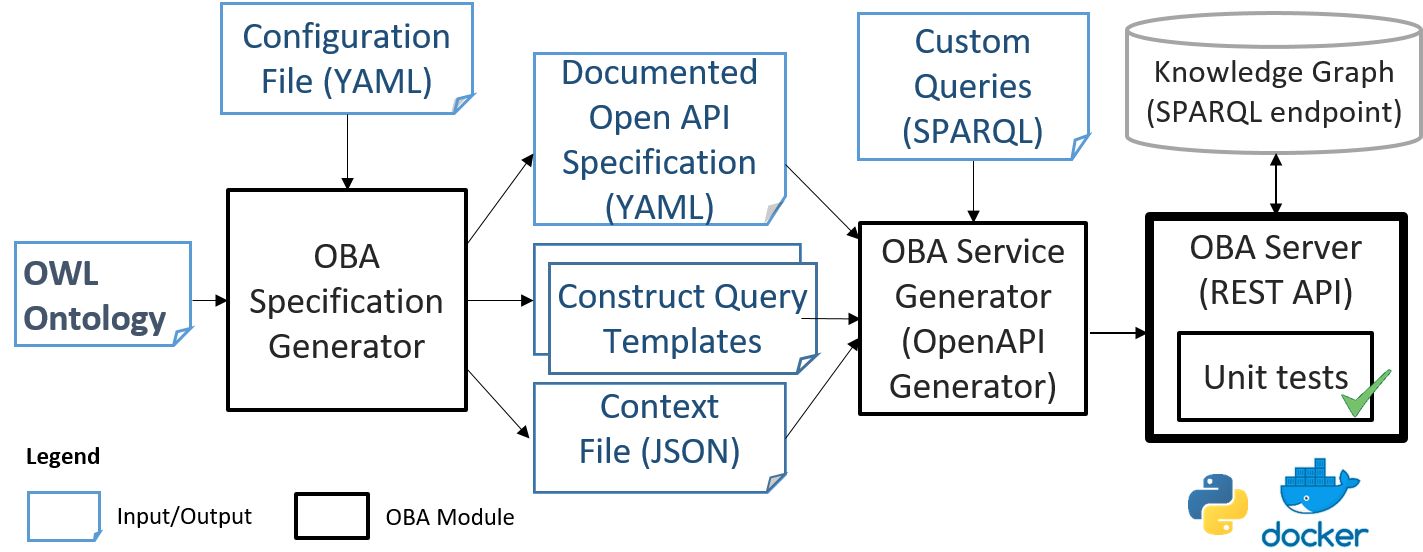}
\caption{Overview of the OBA Framework.}
\label{fig:overview}
\end{figure}

Figure \ref{fig:overview} shows an overview of the workflow followed by the OBA framework, depicting the target input ontology on the left and the resultant REST API on the right. OBA consists of two main modules: the \emph{OBA Specification Generator}, which creates an API specification template from an input ontology; and the \emph{OBA Service Generator}, which produces a server with a REST API for a target SPARQL endpoint. In this section we describe the different features of OBA for each module, along with the main design decisions and assumptions adopted for configuring the server.

\subsection{OBA Specification Generator}\label{sec:spec}

One of the drivers for the development of OBA was the need to use standards and file formats commonly used by web developers (who may not necessarily familiar with Semantic Web technologies). Hence, we decided to use the OpenAPI specification\footnote{\url{https://github.com/OAI/OpenAPI-Specification}} for representing REST APIs and JSON as the main interchange file format.

There are three reasons why we chose the OpenAPI specification (OAS): First, it ``defines a standard, programming language-agnostic interface description for REST APIs, which allows both humans and computers to discover and understand the capabilities of a service without requiring access to source code, additional documentation, or inspection of network traffic"\footnote{\url{http://spec.openapis.org/oas/v3.0.3}}. Second, OAS is an open source initiative backed up by industry and widely used by the developer community, with more than 17.000 stars in GitHub and over 6.000 forks. Finally, by adopting OAS we gain access to a wide range of tools\footnote{\url{https://github.com/OAI/OpenAPI-Specification/blob/master/IMPLEMENTATIONS.md}} that can be leveraged and extended (e.g., for generating a server) and are available in multiple programming languages. 

\subsubsection{Generating an OAS from OWL}\label{sec:mapping}~\\

\noindent OAS describes how to define \emph{operations} (GET, POST, PUT, DELETE) and \emph{paths} (i.e., the different API calls) to be supported by a REST API; together with the information about the \emph{schemas} that define the structure of the objects to be returned by each call. OAS also describes how to provide examples, documentation and customization through parameters for each of the paths declared in an API. 

Typically, an OAS would have two paths for each GET operation; and one for POST, PUT and DELETE operations. For instance, let us consider a simple REST API for registering and returning \emph{regions} around the world. An OAS would have the paths `\verb+/regions+' (for returning all available regions) and `\verb+/regions/{id}+' (for returning the information about a region in particular) for the GET operation; the `\verb+/regions+' path for POST;\footnote{Alternatively, `\texttt{/regions/{id}}' may be used to allow developers to use their own ids} and the `\verb+/regions/{id}+' path for PUT and DELETE operations .

In OAS, the \emph{schema} to be followed by an object in an operation is described through its properties. For example, we can define a \emph{Region} as a simple object with a \emph{label}, a \emph{type} and a \emph{partOfRegion} property which indicates that a region is part of another region.  The associated schema would look as follows in OAS:

{\footnotesize
\begin{Verbatim}
     Region:
      description: A region refers to an extensive, continuous 
            part of a surface or body.
      properties:
        id:
          nullable: false
          type: string
        partOfRegion:
          description: Region where the region is included in.
          items:
            $ref: '#/components/schemas/Region'
          nullable: true
          type: array
        label:
          description: Human readable description of the resource
          items:
            type: string
          nullable: true
          type: array
        type:
          description: type of the resource
          items:
            type: string
          nullable: true
          type: array
      type: object
\end{Verbatim}
}

Note that the \emph{partOfRegion} property will return objects that follow the \emph{Region} schema (as identified by `\verb+/#/components/schemas/Region+'). The \emph{nullable} parameter indicates that the target property is optional.

The main OAS structure maps naturally to the way classes and properties are specified in ontologies and vocabularies. Therefore, in OBA we map\footnote{We follow the best practices for RESTful API design: paths are in non-capital letters and always in plural (e.g., \texttt{/regions}, \texttt{/persons}, etc.)} each ontology class to a different path in the API specification; and we add each object property and data type property in the target ontology to the corresponding schema by looking into its domain and range (complex class restrictions consisting on multiple unions and intersections are currently not addressed). Documentation for each path and property is included in the \verb$description$ field of the OAS by looking at the available ontology definitions (e.g., \texttt{rdfs:comment} annotations on classes and properties). Unions in property domains are handled by copying the property into the respective class schemas (e.g., if the domain of a property is \texttt{`Person or Cat'}, the property will be added in the \texttt{Person} and \texttt{Cat} schemas); and properties declared in superclasses are propagated to their child class schemas. Properties with no domain or range are by default excluded from the API, although this behavior can be configured in the application. By default, all properties are \emph{nullable} (optional). The full mapping between OAS and OWL supported by OBA is available online.\footnote{\url{https://oba.readthedocs.io/en/latest/mapping/}}

Finally, we also defined two filtering features in OBA when generating the OAS to help interacting with the API. First, we allow specifying a subset of classes of interest to include in an API, since ontologies may contain more classes than the ones we may be interested in. Second, by default OBA will define a parameter on each GET path to allow retrieving entities of a class based on their label.  

As a result of executing the OBA specification generator, we create an OAS in YAML format\footnote{\url{https://yaml.org/}} that can be inspected by ontology engineers manually or using an online editor.\footnote{\url{https://editor.swagger.io/}} This specification can be implemented with the OBA server (described in Section \ref{ref:server}) or by alternative means (e.g., by implementing the API by hand).

\subsubsection{Generating SPARQL Query Templates and JSON-LD Context}~\\

\noindent The OBA Specification Generator also creates a series of templates with the queries to be supported by each API path. These queries will be used by the server for automatically handling the API calls. For example, the following query is used to return all the information of a resource by its id (\verb+?_resource_iri+):

{\footnotesize
\begin{Verbatim}
#+ summary: Return resource information by its resource_iri
PREFIX rdfs: <http://www.w3.org/2000/01/rdf-schema#>
CONSTRUCT {
    ?_resource_iri ?predicate ?prop .
    ?prop a ?type .
    ?prop rdfs:label ?label
}
WHERE {
    ?_resource_iri ?predicate ?prop
    OPTIONAL {
        ?prop  a ?type
        OPTIONAL {
            ?prop rdfs:label ?label
        }
    }
}
\end{Verbatim}
}

The individual id \verb+?_resource_iri+ acts as a placeholder which is replaced with the URI associated with the target path (we reuse GRLC \cite{sack_grlc_2016} to define parameters in a query\footnote{\url{https://github.com/KnowledgeCaptureAndDiscovery/OBA_sparql/}}). Returned objects will have one level of depth within the graph (i.e., all the outgoing properties of a resource), in order to avoid returning very complex objects. This is useful in large knowledge graphs such as DBpedia \cite{dbpedia}, where returning all the sub-resources included within an instance may be too costly. However, this default behavior can be customized by editing the proposed resource templates or by adding a custom query (further explained in the next section). 

Together with the query templates, OBA will generate a JSON-LD context file from the ontology, which will be used by the server to translate the obtained results back into JSON. We have adapted owl2jsonld \cite{soiland-reyes_owl2jsonld_2014} for this purpose.

\subsection{OBA Service Generator}\label{ref:server}
Once the OAS has been generated, OBA creates a script to set up a functional server with the API. We use OpenAPI generator,\footnote{\url{https://github.com/OpenAPITools/openapi-generator}} one of the multiple server implementations for OAS made available by the community; to generate a server with our API as a Docker image.\footnote{\url{https://oba.readthedocs.io/en/latest/server/#execute-server-generation-scripts}} Currently, we support the Python implementation, but the architecture is flexible enough to change the server implementation in case of need. OBA also includes a mechanism for enabling pagination, which allows limiting the number of resources returned by the server.

OBA handles automatically several aspects that need to be taken into account when setting up a server, including how to validate and insert complex resources in the knowledge graph, how to handle authentication; how to generate unit tests and how to ease the access to the server by making clients for developers. We briefly describe these aspects below. 


\subsubsection{Converting SPARQL results into JSON}~\\

\begin{figure}[t!]
\centering
\includegraphics[width=\textwidth]{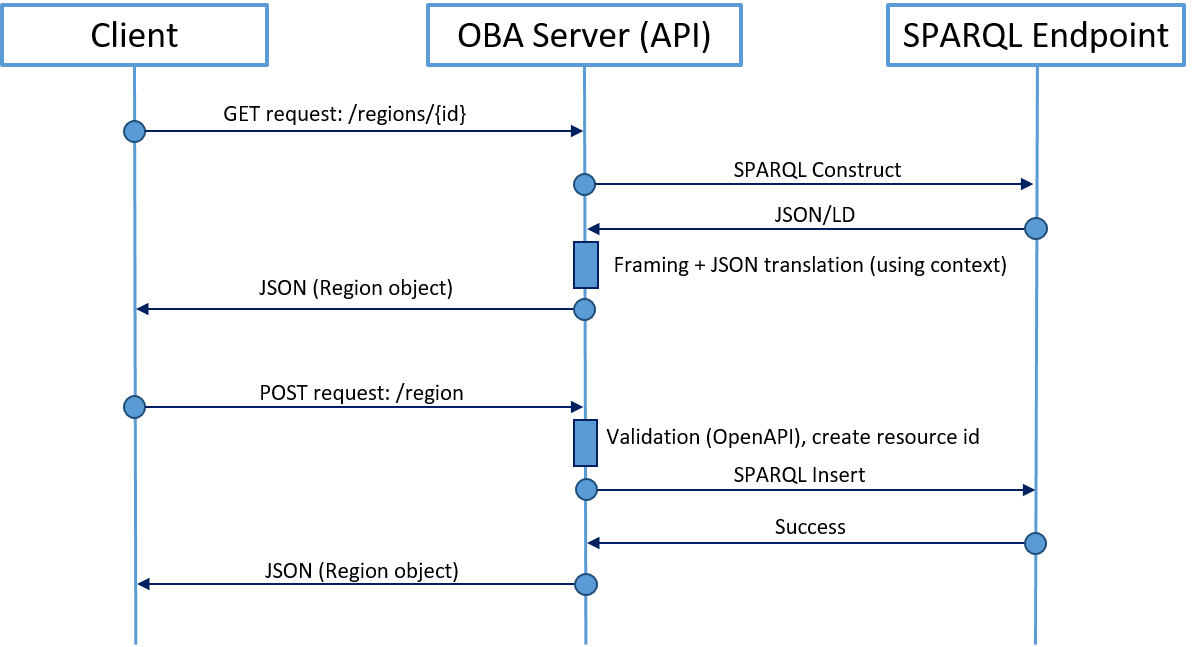}
\caption{Sample GET and POST request through the OBA server.}
\label{fig:sequence}
\end{figure}

\noindent We designed OBA to generate results in JSON, one of the most popular interchange formats used in web development. Figure \ref{fig:sequence} shows a sequence diagram with the steps we follow to produce the target JSON in GET and POST requests. For example, for the GET request, we first create a SPARQL CONSTRUCT query to retrieve the result from a target knowledge graph. The query is created automatically using the templates generated by OBA, parametrizing them with information about the requested path. For example, for a GET request to \verb+/regions/id+, the \verb+id+ will replace   \verb+?_resource_iri+ in the template query as described in Section 3.1.2. 

As shown in Figure \ref{fig:sequence}, the construct query returns a JSON-LD file from the SPARQL endpoint. We frame the results to make sure they follow the structure defined by the API and then we transform the resultant JSON-LD to JSON. In order to transform JSON-LD to JSON and viceversa, we keep a mapping file with the API path to ontology class URI correspondence, which is automatically generated from the ontology. The URI structure followed by the instances is stored in a separate configuration file. 



\subsubsection{Resource Validation and Insertion}~\\

\noindent OBA uses the specification generated from an input ontology to create a server with the target API. By default, the server is prepared to handle GET, POST, PUT and DELETE requests, which are addressed with CONSTRUCT, INSERT, UPDATE and DELETE SPARQL queries respectively. However, POST, PUT and DELETE requests need to be managed carefully, as they modify the contents of the target knowledge graph.

For POST and PUT, one of the main issues to address is handling \emph{complex objects}, i.e., objects that contain one or multiple references to other objects that do not exist in the knowledge graph yet. Following our above example with regions, what would happen if we received a POST request with a new region where \emph{partOfRegion} points to other regions that do not exist yet in our knowledge graph? For example, let us consider that a developer wants to register a new region \texttt{Marina del Rey} that is part of \texttt{Los Angeles}, and none of them exist in the knowledge graph. One way would be requiring the developer to issue a new request to register each parent region before registering the child one (e.g., a POST request first to register \texttt{Los Angeles} region and then another POST request for \texttt{Marina del Rey}); but this makes it cumbersome for developers to use API. Instead, OBA deals with this issue in a recursive manner: If a resource does not have an id, OBA tries to insert all the resources it points to, validates the resource against the corresponding schema and inserts it in the knowledge graph. Hence, in the previous example OBA would register \texttt{Los Angeles} region first, and then \texttt{Marina del Rey}. If a resource already has an id, then it will not be registered as a new resource. When a new resource is created, the server assigns its id with a uuid, and returns it as part of the JSON result. 

This recursive behavior is not desirable for DELETE requests, as we could potentially remove a resource referenced by other resources in the knowledge graph. Therefore, OBA currently deletes only the resource identified by its id in the request.

Finally, OBA defines a simple mechanism for different users to contribute and retrieve information from a knowledge graph. By default, users are assigned a \emph{named graph} in the target knowledge graph. Each named graph allows users submitting contributions and updates independently of each other (or collaboratively, if they share their credentials). User authentication is supported by default through Firebase,\footnote{\url{https://firebase.google.com/docs/auth/}} which leverages standards such as OAUth2.0\footnote{\url{https://oauth.net/2/}} for easy integration. However, we designed authentication to be extensible to other authentication methods, if desired. OBA may also be configured so users can retrieve all information from all available graphs; or just their own graph. User management (i.e., registering new users) is out of the scope of our application.


\subsubsection{Support for Custom Queries}~\\

\noindent OBA defines common template paths from an input ontology, but knowledge engineers may require exposing more complex queries to web developers. For example, knowledge engineers may want to expose advanced filtering (e.g., return regions that start with ``Eu"), have input parameters or complex path patterns (e.g., return only regions that are part of another region). These paths are impossible to predict in advance, as they depend on heterogeneous use cases and requirements. Therefore, in OBA we added a module to allow supporting \emph{custom queries} and let knowledge engineers expand or customize any of the queries OBA supports by default. 

To add a custom query, users need to follow two main steps. First,  create a CONSTRUCT query in a file with the target query; and second,  edit the OAS with the path where the query needs to be supported. OBA reuses GRLC's query module \cite{sack_grlc_2016} to support this feature, as GRLC is an already a well established application for creating APIs from SPARQL queries. An example illustrating how to add custom queries to an OAS in OBA can be found online.\footnote{\url{\url{https://oba.readthedocs.io/en/latest/adding\_custom\_queries/}}}   


\subsubsection{Generating Unit Tests}~\\

\noindent OBA automatically generates unit tests for all the paths specified in the generated OAS (using Flask-Testing,\footnote{\url{https://pythonhosted.org/Flask-Testing/}} a unit test toolkit for Flask servers\footnote{\url{https://flask.palletsprojects.com/en/1.1.x/}} in Python). Units tests are useful to check if the data in a knowledge graph is consistent with the classes and properties used in the ontology, and to identify unused classes. By default, OBA supports unit tests for GET requests only, since POST, PUT and DELETE resources may need additional insight of the contents stored in the target knowledge graph. However, this is a good starting point to test the different API calls to be supported to the API and detect any inconsistencies. Knowledge engineers may extend the test files with additional tests required by their use cases. Unit tests are generated as part of the server, and may invoked before starting up the API for public consumption.\footnote{\url{https://oba.readthedocs.io/en/latest/test/}} 


\subsubsection{Generating Clients for API Exploitation}~\\

\noindent The OpenAPI community has developed tools to generate clients  to support an OAS in different languages. We use the OpenAPI generator in OBA to create clients (software packages) to ease API management calls for developers. For example, below is an example of a code snippet using the Python client for one of the APIs we generated with OBA\footnote{\url{https://model-catalog-python-api-client.readthedocs.io/en/latest/}} and available through \texttt{pip}. The client retrieves the information of a region (with id `\emph{Texas}') and returns the result as a JSON object in a python dictionary without the need of issuing GET requests or writing SPARQL:

{\footnotesize
\begin{Verbatim}
import modelcatalog 
# modelcatalog is the python package with our API
api_instance = modelcatalog.ModelApi()
region_id = "Texas"
try:
    # Get a Region by its id
    region = api_instance.regions_id_get(region_id)
    print(region) 
    # Result is a JSON object with the region properties in a dictionary
except ApiException as e:
    print("Exception when calling ModelApi->regions_id_get: %s\n" % e)
\end{Verbatim}
}

\section{Showcasing OBA's Features through Examples}\label{sec:showcase}
In this section we demonstrate the full capabilities of OBA in an incremental manner through three different examples of increasing complexity. All the OBA configuration files required to generate the OAS described in this section are accessible online.\footnote{\url{https://oba.readthedocs.io/en/latest/examples/}} We describe these examples below:


\paragraph{Drafting an API for an ontology network:} The simplest way in which OBA can be used is by generating a draft OAS from a given ontology (without generating a server). We have tested OBA with ten different ontologies\footnote{\url{https://github.com/KnowledgeCaptureAndDiscovery/OBA/tree/master/examples}} from different domains to generate draft specifications, and we have found this feature very useful in our work. Drafting the API allows knowledge engineers discuss potential errors for misinterpretation, as well as easily detect errors on domains and ranges of properties.

\paragraph{Generating a GET API for a large ontology:} DBPedia \cite{dbpedia} is a popular knowledge graphs with millions of instances over a wide range of categories. The DBPedia ontology\footnote{\url{https://wiki.dbpedia.org/services-resources/ontology}} contains over 680 classes and 2700 properties; and creating an API manually to support them becomes a time consuming task. 
We demonstrated OBA by creating two different APIs for DBPedia. The first API contains all the paths associated with the classes in the ontology, showing how OBA can be used by default to generate an API even when the ontology has a considerable size. Since the resultant API is too big to browse manually, we created a Python client\footnote{\url{https://github.com/sirspock/dbpedia_api_client}} and a notebook\footnote{\url{https://github.com/sirspock/dbpedia_example/blob/master/scientists_get.ipynb}} demonstrating its use. The second API has just a selected group of classes by using a filter, as in some cases not all the classes may need to be supported in the desired API. OBA does a transitive closure on the elements that are needed as part of the API. For example, if the filter only contains ``Band", and it has a relationship with ``Country" (e.g., \emph{origin}), then by default OBA will import the schema for ``Country" into the API specification to be validated accordingly. In the DBPedia example, selecting just 2 classes (\verb+dbpedia:Genre+ and \verb+dbpedia:Band+) led to the inclusion of more than 90 paths in the final specification. 



\paragraph{Generating a full Create, Delete, Update, Delete (CRUD) API:} OKG-Soft \cite{garijo_okg-soft_2019} is an open knowledge graph with scientific software metadata, developed to ease the understanding and execution of complex environmental models (e.g., in hydrology, agriculture or climate sciences). A key requirement of OKG-Soft was for users to be able to contribute with their own metadata in collaborative manner, and hence we used the full capabilities of OBA to support adding, editing and deleting individual resources. OKG-Soft uses two ontologies to structure the knowledge graph, which have evolved over time with new requirements. We used OBA to maintain an API release after each ontology version, generating an OAS, updating it with any required custom queries and generating a server with unit tests, which we executed before deploying the API in production. Having unit tests helped detecting and fixing inconsistencies in the RDF, and improved the overall quality of the knowledge graph. Authenticated users may use the API for POST, PUT and DELETE resources;\footnote{\url{https://model-catalog-python-api-client.readthedocs.io/en/latest/endpoints/}} and we use the contents of the knowledge graph for model exploration, setup and execution in different environments. An application for browsing the contents of the knowledge graph is available online.\footnote{\url{https://models.mint.isi.edu}}


The three examples described in this section demonstrate the different features of OBA for different ontologies: the ability to draft API specifications, the capabilities of the tool to be used for large ontologies and to filter classes when required; and the support for GET, POST, PUT and DELETE operations while following the best practices for RESTful design. 




\section{Adoption, Impact and Limitations}\label{sec:adoption}
\begin{figure}[t!]
\centering
\includegraphics[width=\textwidth]{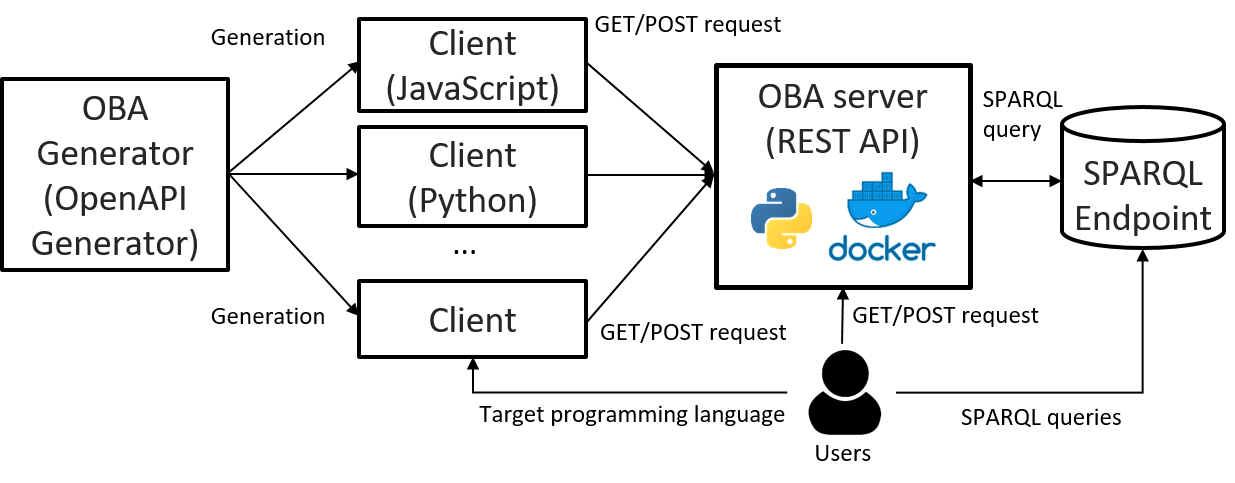}
\caption{OBA: Client side architecture}
\label{fig:client}
\end{figure}

We developed OBA to help developers (not familiar with SPARQL) accessing the contents of knowledge graphs structured by ontologies. Figure \ref{fig:client} shows an overview of the different ways that OBA supports user access to the contents of a knowledge graph (using a SPARQL endpoint). Non-expert web developers may use clients in the languages they are more familiar with (e.g., Python, JavaScript, etc.); generated with the OBA Service Generator. Web developers with more knowledge on using APIs may use the API created with the OBA server. Knowledge engineers may choose to query the SPARQL endpoint directly.

OBA builds on the work started by tools like Basil \cite{basil} and GRLC \cite{sack_grlc_2016} -pioneers in exposing SPARQL queries as APIs- to help involve knowledge engineers in the process of data retrieval from their knowledge graphs with their ontologies. In our experience, generating a draft API from an ontology has helped our developer collaborators understand how to consume the information of our knowledge graphs, while helping the ontology engineers in our team detect potential problems in the ontology design. 

In fact, similar issues have been raised in the Semantic Web community for some time. For example, the lack of guidance when exploring existing SPARQL endpoints\footnote{\url{https://lists.w3.org/Archives/Public/semantic-web/2015Jan/0087.html}} has led to the development of tools such as \cite{DBLP:conf/semweb/Mihindukulasooriya15} to help finding patterns in SPARQL endpoints in order to explore them. The Semantic Web community has also acknowledged the difficulties developers experience to adopt RDF,\footnote{\url{https://lists.w3.org/Archives/Public/semantic-web/2018Nov/0036.html}} which have resulted in ongoing efforts to improve materials and introductory tutorials.\footnote{\url{https://github.com/dbooth-boston/EasierRDF}} 

We believe OBA helps addressing these problems by exploiting Semantic Web technologies while exposing the information to developers following the REST standards they familiar with. OBA allows prototyping APIs from ontologies, helps maintainability of the APIs (having an API per version of the ontology), helps validation of the API paths and contents of the knowledge graph assisting in the creation of unit tests and includes documentation for all of the API schemas automatically. In addition, the tool is thoroughly documented, with usage tutorials and examples available online.\footnote{\url{https://oba.readthedocs.io/en/latest/quickstart/}}




We end this section by discussing assumptions and limitations in OBA. For instance, OBA assumes that the target endpoint is modeled according to the ontology used to create the API; and changes in the ontology version will lead to a new version of the API (hence keeping track of which version supports which operations). OBA also assumes that two classes in an ontology network don't have the same local name, as each class is assigned a unique path. As per current limitations, OBA simplifies some restrictions in the ontology, such as complex axioms in property domains and ranges (e.g., having unions and intersections at the same time as a property range), to help creating the OAS. In addition, for large ontologies the resultant APIs may be large, which will work appropriately handling requests, but may be slow to render in a browser (e.g., to see documentation of a path). OBA is proposed as a new resource, and therefore we don't have usage metrics from the community so far.

\section{Conclusions and Future Work}\label{sec:conclusions}

In this paper we have introduced the Ontology-Based APIs framework (OBA), a new resource for creating APIs from ontologies by using the OpenAPI Specification. OBA has demonstrated to be extremely useful in our work, helping setting up and maintaining API versions, testing and easy prototyping against a target knowledge graph. We believe that OBA helps bridging the knowledge gap between  ontology engineers and developers, as it provides the means to create a guide (a documented API) that illustrates how to exploit a knowledge graph using the tools and standards developers are used to.

We are actively expanding OBA to support new features. First, we are working towards supporting additional mappings between OWL and OAS, such as complex domain and range axiomatization. Second, we are working to support accepting and delivering JSON-LD requests (instead of JSON only), which is preferred by some Semantic Web developers. As for future work, we are exploring the possibility of adding support for GraphQL, which has gained popularity lately, as an alternative to using SPARQL to retrieve and return contents. Finally, an interesting approach worth exploring is to use combine an ontology with existing tools to mine patterns from knowledge graphs to expose APIs with the most common data patterns. 

\section*{Acknowledgements}
This work was funded by the Defense Advanced Research Projects Agency with award W911NF-18-1-0027 and the National Science Foundation with award ICER-1440323. The authors would like to thank Yolanda Gil, Paola Espinoza, Carlos Badenes and Oscar Corcho for their thoughtful comments and feedback.










\bibliographystyle{splncs04}

\bibliography{bib}
\end{document}